\documentclass[10pt,journal]{IEEEtranTCOM}
\usepackage[english]{babel}
\usepackage[usenames]{color}
\usepackage[cp1250]{inputenc}
\usepackage{amsfonts}
\usepackage{amsthm}
\usepackage{graphicx}
\usepackage{epsfig}
\usepackage{mathrsfs}
\usepackage{amsmath}
\usepackage{algorithm}
\usepackage{algorithmic}

\theoremstyle{plain}

\begin{document}
\newcommand{\bea}{\begin{eqnarray}}
\newcommand{\eea}{\end{eqnarray}}
\newcommand{\be}{\begin{equation}}
\newcommand{\ee}{\end{equation}}
\newcommand{\beas}{\begin{eqnarray*}}
\newcommand{\eeas}{\end{eqnarray*}}
\newcommand{\bs}{\backslash}
\newcommand{\bc}{\begin{center}}
\newcommand{\ec}{\end{center}}
\def\SC {\mathscr{C}}

\title{Credibility evaluation of income data\\ with hierarchical correlation reconstruction }
\author{\IEEEauthorblockN{Jaros\l aw Duda$^{\star}$ \qquad Adam  Szulc$^{\dagger}$}\\
\IEEEauthorblockA{$^{\star}$ {Jagiellonian University, Cracow, Poland, Email: dudaj@interia.pl}\\
   $^{\dagger}$ {Warsaw School of Economics, Warsaw, Poland}}}
\maketitle

\begin{abstract}
In situations like tax declarations or analyzes of household budgets we would like to automatically evaluate credibility of exogenous variable (declared income) based on some available (endogenous) variables - we want to build a model and train it on provided data sample to predict (conditional) probability distribution of exogenous variable based on values of endogenous variables. Using Polish household budget survey data there will be discussed simple and systematic adaptation of hierarchical correlation reconstruction (HCR) technique for this purpose, which allows to combine interpretability of statistics with modelling of complex densities like in machine learning. For credibility evaluation we normalize marginal distribution of predicted variable to $\rho\approx 1$ uniform distribution on $[0,1]$ using empirical distribution function $(x=EDF(y)\in[0,1])$, then model density of its conditional distribution $(\textrm{Pr}(x_0|x_1 x_2\ldots))$ as a linear combination of orthonormal polynomials using coefficients modelled as linear combinations of features of the remaining variables. These coefficients can be calculated independently, have similar interpretation as cumulants, additionally allowing to directly reconstruct probability distribution. Values corresponding to high predicted density can be considered as credible, while low density suggests disagreement with statistics of data sample, for example to mark for manual verification a chosen percentage of data points evaluated as the least credible.
\end{abstract}
%\textbf{Keywords:} credibility evaluation, statistics, machine learning, predicting probability distribution
\section{Introduction}
While in standard regression we want to estimate the conditional expected value, in some situations we need to predict the entire probability distribution. For example in ARMA/ARCH modelling~\cite{arma}, or data compression like JPEG-LS~\cite{loco}, it is resolved by predicting the most likely value, usually using a linear combination of neighboring values, then assuming some unimodal parametric distribution of error from this prediction - for example as Gaussian in ARMA-like models or usually Laplace distribution in data compression. Widths of such distributions are often chosen based on the context (e.g. in ARCH, JPEG-LS).

However, in situation like credibility evaluation of tax declarations, the conditional distribution of the main variable to verify (declared income) is quite complex and often multimodal, as we will see in the presented analysis (e.g. Fig. \ref{cont}), and we need to model dependencies with multiple types of variables - their features might affect not only the expected value, but also higher moments. Such dependence between expected values of two variables is described by correlation coefficient, between variances is considered e.g. in ARCH model, HCR used here allows to systematically exploit such dependencies of any moments between two or more variables.

\begin{figure}[t!]
    \centering
        \includegraphics{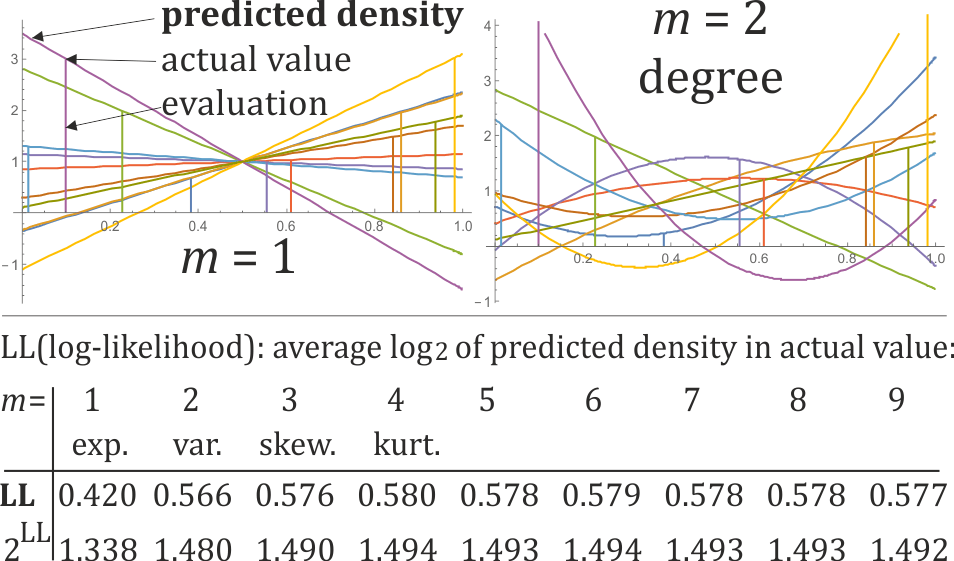}
        \caption{Top: modelled conditional densities of $\textrm{Pr}(x_0|x_1\ldots x_d)$ as degree $m=1$ (left) or $m=2$ (right) polynomials (Fig. \ref{poly} contains further up to $m=10$, calibration to interpret as nonnegative and translation $x_0\to y_0$ to density of original variable) for predicted exogenous variable (equivalent income normalized to uniform marginal distribution on $[0,1]$) based on the remaining (endogenous) variables for ten randomly chosen data points from the sample. The actual values are marked with vertical lines, the higher predicted density for them, the better - we see that in most cases the prediction is above the base $\rho=1$. Inconvenience of this method is sometimes obtaining negative densities, calibrated to nonnegative in Fig. \ref{poly}, here we can interpret such predictions as just having low credibility. Bottom: evaluation of such prediction using log-likelihood (LL). There was randomly chosen 75\% of points from dataset to train the model: calculate coefficients, the remaining 25\% were used for evaluation: average $\log_2$ of predicted normalized (for $x_0\in [0,1]$) conditional density in the actual value. The table shows averages of 10 such experiments for $m=1,\ldots,9$ degree of polynomial for predicted density, the size of model is $223m$ coefficients. We see that $m=4$ seems a reasonable compromise (also later in Fig. \ref{expvar}), can be interpreted as using moments up to kurtosis, all its 892 coefficients with interpretations are presented in Fig. \ref{coef}. Log-likelihood corresponds to average increase of predicted density, for example presented $m=2$ predicted densities already give on average $\approx 1.5$ larger density for actual points than the base $\rho=1$.         }
       \label{intr}
\end{figure}

To evaluate credibility of a given (exogenous) variable $y_0$, we would like to predict its conditional probability density of $\textrm{Pr}(y_0|y_1\ldots y_d)$ based on the remaining (endogenous) variables $y_1\ldots y_d$ - intuitively, the higher such density is, the higher credibility of $y_0$. However, this intuition requires some normalization, as for example tails of distributions have lower density, but it should not be interpreted as lower credibility - these values are just spread over wider range.

A natural normalization, used for example in copula theory~\cite{copula}, is to nearly uniform marginal distribution $\rho_0(x_0)\approx 1$ on $[0,1]$. We can use empirical distribution function (EDF, by sorting obtained values) to transform the original variable $y$ to $x=EDF(y)$ from nearly uniform distribution. Using this normalization, modelled $\rho_0(x_0)$ density of $\textrm{Pr}(x_0|x_1\ldots x_d)$ can be seen as evaluation of credibility, its examples are presented in Fig. \ref{intr}.

To model $\textrm{Pr}(x_0|x_1\ldots x_d)$ we will use hierarchical correlation reconstruction (HCR) approach~(\cite{me1,me2,me3,me4}): model density of joint distribution of all variables $[0,1]^{d+1}$ as a linear combination of orthonormal polynomials:
\be \rho(\textbf{x}) =\sum_{\textbf{j}\,=\,(j_0\ldots j_d)} a_{\textbf{j}}\, f_{j_0}(x_0)\cdot \ldots\cdot f_{j_d} (x_d)\ee
where $(f_j)_j$ satisfy $\int_0^1 f_i(x) f_j(x) dx = \delta_{ij}$.

Coefficients $a_\textbf{j}$ are mean-square error (MSE) optimized: minimizing $L_2$ norm between polynomial and Gaussian-smoothened sample in width$\to 0$ limit (to Dirac deltas). It uses proper weights thanks to normalization to uniform in $[0,1]$: proportional to percentage of population.

These $a_{\textbf{j}}$ coefficients have cumulant-like interpretation. For example $a_{100\ldots 0}$ has similar behavior as the expected value of the first variable, $a_{020\ldots 0}$ as variance of the second variable. We can also use mixed coefficients this way, for example $a_{110\ldots 0}$ describes dependence between expected values of the first two variables - has similar interpretation as correlation coefficient. We can model complex joint density with such polynomial approximation, especially that MSE estimation of these coefficient turns out very inexpensive and can be calculated independently~\cite{me1}: using orthonoromal basis, coefficient of a given function turns out just average of this function over the sample: $a_{\textbf{j}}=\frac{1}{n} \sum_{1=1}^n f_{\textbf{j}}(\textbf{x}^i)$ for $\textbf{x}^i=(x^i_0,\ldots,x^i_d)$ data points.\\

For credibility evaluation, after normalizing all marginal distributions to nearly uniform on $[0,1]$, we would like to estimate their joint density as such linear combination - using a chosen basis, for example exploiting only pairwise dependencies (up to two nonzero indexes). Then substituting $x_1\ldots x_d$ coordinates to such joint distribution, and normalizing to integrate to 1, we get probability density $\rho_0(x_0)$, describing evaluated credibility of this value.

However, above HCR approach is appropriate for continuous variables, while here we have many discrete - there will be suggested adaptation for this situation. Finally, for simplicity and interpretability there will be just used least-squares regression to optimize linear dependencies between features of endogenous variables $(v)$ and cumulant-like coefficients of the predicted variable:
$$\rho_0(x_0)=1+\sum_j a_j f_j(x_0)\quad\textrm{for}\quad a_j=\beta_0^j +\sum_k \beta_k^j v_k$$
with coefficients chosen by least-square optimization of $\sum_i\| f_j(x_0^i) -a_j\|^2$.

Its evaluation is presented in Fig. \ref{intr}: by calibrating predicted densities to nonnegative and calculating log-likelihood: average $\log_2$ of predicted density in actual value. To avoid overfitting, the coefficients are calculated using randomly chosen 75\% of points, evaluation is made on the remaining 25\%. Such procedure was repeated 10 times and its averages are presented, $m=4$ seems the best compromise.

Calibration is not required for credibility evaluation. We can  just directly calculate values of predicted polynomials, and e.g. choose a threshold below which we can recommend some further action like additional verification. For example choosing this threshold value as just 0, it would mark least credible looking $\approx 1\%$ of population for further verification.

%For its evaluation/calibration there will be used plot of sorted $\rho_0(x_0)$ for all $n$ points from the dataset, presented in Fig. \ref{intr}. Its horizontal axis can be seen as quantile regarding modelled credibility. For example choosing that we want to manually verify 1\% of the least credible data points, we can read the density threshold ($\rho\approx 0$) from this plot and verify only those below this threshold.

\section{Dataset}
\begin{figure}[b!]
    \centering
        \includegraphics{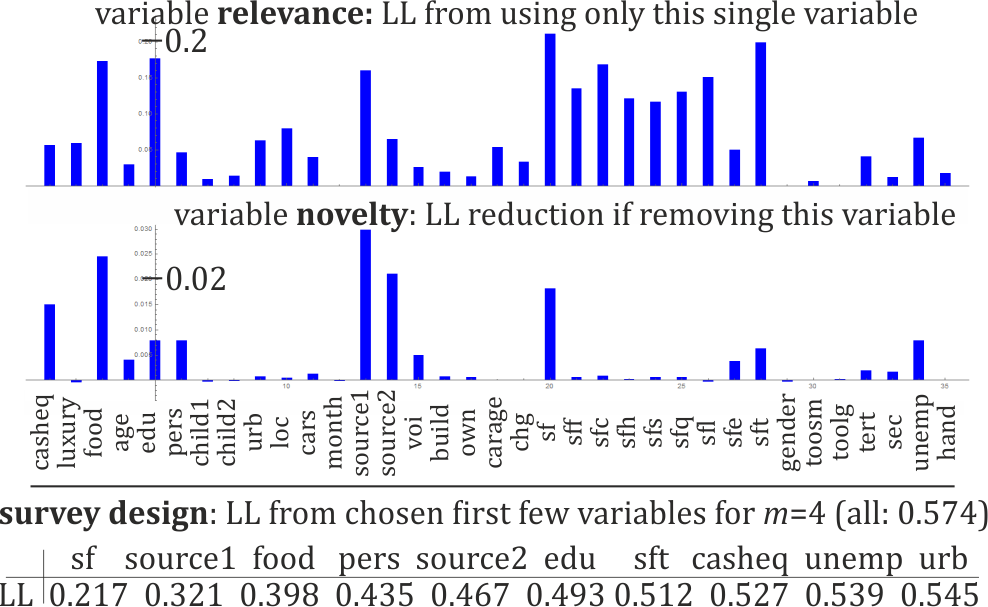}
        \caption{Top: evaluation of importance of used variables. Relevance as log-likelihood of prediction using only a given single variable, strength of statistical dependence between the two variables (can be interpreted as estimated $-h(X_0|X_i)$ conditional entropy). Novelty shows reduction of log-likelihood if removing given variable from the set of all of them, low novelty means that given variable is redundant here: its contribution can be inferred from the remaining variables. Bottom: log-likelihood from a few variables up to a given position in the table, for example using first 5 variables we get 0.467 log-likelihood. Their order was optimized in a greedy way (not necessarily optimal) - succeedingly choosing single new variables maximizing contribution.        }
       \label{var}
\end{figure}
The dataset is composed of observations on 37215 households collected annually by the Central Statistical Office of Poland (GUS). There will be used the following $36$ variables (all but the first are endogenous - used to predict the first variable which is exogenous; further used short variable names are provided in brackets):
\begin{enumerate}
  \item Continuous: equivalent monthly income\footnote{It is calculated as the household disposable income divided by a respective equivalence scale i. e. an indicator supposed to measure impact of demographic variables on cost of living.  In the present study simple OECD 70/50 formula is employed: scale = 1 + (number of adults minus one) $\cdot$ 0.7 + (number of children) $\cdot$ 0.5.} (inceq) in PLN - the exogenous variable, remaining equivalent cash in PLN at the end of month (casheq), shares of expenditures on luxury goods and food (luxury and food, respectively).
  \item Discrete ordinal (the number of distinct values is provided in brackets): age of the household head in years (age, 87 values), his/her completed education level (edu, 11), number of persons (pers, 14), number of younger (child1, 8) and older (child2, 7) children in the household, urbanisation level (urb, 3), type of residence (loc, 6), number of cars (cars, 7), month of the query (month, 12).
  \item Discrete categorical: main income source (source1, 12) and additional source (source2, 13), voivodship (voi, 16), building type (build, 4), its ownership type (own, 6), age of the newest car (carage, 5), subjective evaluations of: change in the material position (chg, 5), income sufficiency (sf, 5), level of satisfaction of needs for food (sff, 5), clothing (sfc, 5), health care (sfh, 6), housing fees (sfs, 6), housing equipment (sfq, 6), culture (sfl, 7), education (sfe, 6), tourism and recreation (sft, 6).
  \item Binary: gender of the household head (gender), subjective evaluation whether the dwelling is too small or too large (toosm, toolg), presence in the household of persons: with tertiary and secondary education (tert, sec), unemployed (unemp), handicapped (hand).
\end{enumerate}

The model will use linear combinations of their features. Age will be treated as continuous variable, pairwise dependencies for all four can be seen in Fig. \ref{cont}, their cumulant-like proprieties are used in linear combination for the prediction. To avoid arbitrary choice of weights, all the remaining variables are treated as binary: split into 0/1 variables, as many as the number of distinct values, being 1 if the category agrees, 0 otherwise (called one hot encoding in machine learning). Final coefficients are presented in Fig. \ref{coef}.

\section{Normalization and orthogonal basis}
For many reasons it is convenient to normalize variables to have nearly uniform marginal distributions ($y\to x$): to see them as quantiles - interpreting ranges as population percentages hence getting proper weights, to directly interpret predicted density as credibility here, finally to use polynomials for density estimation with normalized coefficients. In previous applications of HCR~(\cite{me3,me4}), there was used CDF (cumulant distribution function) of approximated parametric distribution (Laplace) for this normalization - approximating general behavior, then modelling  corrections from this idealization with polynomial, which can evolve in time for non-stationary time series. Here probability distribution is stationary and seems too complex for parametric distributions, hence, like in copula theory~\cite{copula}, we will directly use EDF for this normalization.

\subsection{Normalization with empirical distribution function}
\begin{figure}[t!]
    \centering
        \includegraphics{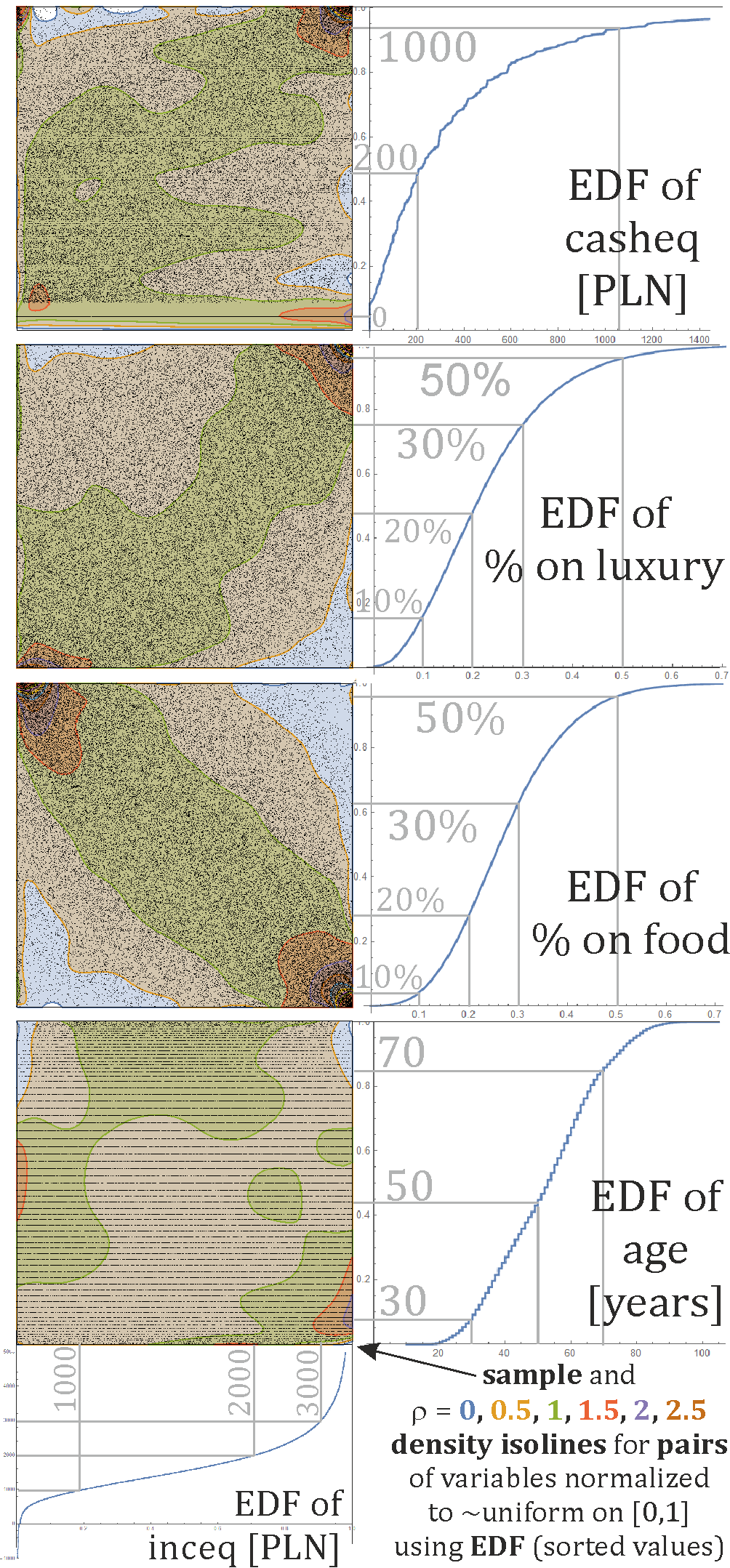}
        \caption{Pairwise dependencies between 5 variables treated as continuous: exogenous (equivalent income) on horizontal axis to be predicted from endogenous variables on vertical axes. Each is normalized to nearly uniform marginal distribution - position can be seen as quantile, 0.5 as median, length e.g. 0.2 as 20\% of population. Some of them have discreteness - corresponding to horizontal dotted lines. Each of four $[0,1]^2$ presented diagrams contains 37215 black dots from the dataset, and isolines for their density (would be $\rho\approx 1$ for independent variables) - estimated with HCR as polynomial $\sum_{ij=0}^9 a_{ij} f_i (x_0) f_j(x_1)$ using 100 coefficients (mixed moments up to 9th). For example for age we can see that younger people have higher expected inceq, middle-age lower, older closer to median (lower variance) - we need at least second order polynomial $(f_2)$ of age to model such behavior. }
       \label{cont}
\end{figure}
Having $y^1,\ldots,y^n$ sample, we can normalize it with EDF by sorting the values - finding order (bijection) $o:\{1\ldots n\}\to\{1\ldots n\}$ such that:
\be y^{o(1)}\leq y^{o(2)}\leq \ldots \leq y^{o(n)} \label{sort} \ee
Hence, $y^i$ is in $o^{-1}(i)$-th position of this order - wanting them to have nearly uniform distribution on $[0,1]$, a natural choice is $x^i=\frac{1}{n}(o^{-1}(i)-1/2)$.

However, especially for discrete variables, (\ref{sort}) can have many equalities, what needs a special treatment - there is no base to choose an order among equal values, all of them should be transformed into the same value $x^i$, naturally chosen as the center of such range - we can see one such horizontal line for casheq = 0 in Fig. \ref{cont}, and age sample consisting only of horizontal lines due to rounding to complete years.

Finally, the used generalized formula (working for both continuous and discrete variables) is:
\be x^i=\frac{\min \{k:y^{o(k)}=y^i\}+\max \{k:y^{o(k)}=y^i\}-1}{2n} \label{norm}\ee

We can use this formula to normalize each variable of $\textbf{y}^i=(y^i_0,\ldots,y^i_d)$: separately for each variable (lower index), getting $\textbf{x}^i=(x_0^i,\ldots,x^i_d)$ for continuous variables having nearly uniform marginal distributions on $[0,1]$. However, for discrete variables it is a step function, especially for binary - needing a completely different treatment.

There are ways to transform discrete variables into a smaller number of nearly independent continuous variables - for example by choosing a set of orthonormal projections. This choice is sometimes done randomly, but it often can be optimized by dimensionality reduction techniques like PCA (principal component analysis) or its discrete analogues like MCA (multiple correspondence analysis)~\cite{mca}. Such projections often have nearly Gaussian distribution, but we would loose interpretability this way - this article presents analysis directly using discrete variables to get better interpretation of obtained coefficients (Fig. \ref{coef}), however, it might be worth to explore also methods like MCA to directly work on continuous variables.

\subsection{Orthogonal polynomial basis and HCR}
Assuming (normalized) variable is from nearly uniform distributions on $[0,1]$, it is very convenient to represent its density with polynomial: $\rho(x)=\sum_j a_j f_j(x)$. Using orthonormal basis $\int_0^1 f_i(x) f_j(x)\, dx = \delta_{ij}$, mean-square optimization leads to~\cite{me1} inexpensive estimation by just averaging: $a_j =\frac{1}{n}\sum_{i=1}^n f_j(x^i)$.

The first five of these polynomials (rescaled Legendre) are $f_0=1$, and $f_1,f_2,f_3,f_4$ correspondingly (drawn in Fig. \ref{coef}):
$$\sqrt{3}(2x-1), \sqrt{5}(6x^2-6x+1), \sqrt{7}(20x^3-30x^2+12x-1),$$
$$3(70x^4-140x^3+90x^2-20x+1). $$

Similar alternative natural choice is cosine basis: $\{1,\sqrt{2}\cos(\pi x),\sqrt{2}\cos(2\pi x),\sqrt{2}\cos(3\pi x),\ldots\}$.

As $\int_0^1 f_j(x)dx=0$ for $j\geq 1$, density normalization needs $a_0=1$. The $a_1$ term shifts the expected value toward left or right. Positive $a_2$ increases probability of extreme values - has analogous behavior as variance. And so on: $a_j$ coefficient has similar interpretation as $j$-th cumulant. Using degree $m$ polynomial: $j=0\ldots m$ corresponds to modelling distribution using the first $m$ moments, additionally directly getting density estimation from them.

We can also exploit statistical dependencies between two or more variables this way - by analogously modelling joint distribution on $[0,1]^{d+1}$ using product basis: $\rho(\textbf{x}) =\sum_{\textbf{j}\,=\,j_0\ldots j_d} a_{\textbf{j}}\, f_{j_0}(x_0)\cdot \ldots\cdot f_{j_d} (x_d)$. This way $a_{\textbf{j}}$ represents mixed cumulants - their dependencies between multiple variables. For a large number of variables, most of coordinates of used $\textbf{j}$ should be 0 - coefficients with single nonzero coordinate describe probability distribution of corresponding variable, with two nonzero describe pairwise dependencies and so on - getting hierarchical correlation reconstruction (HCR) of a given distribution.

We could directly use such modelled joint density $\rho(x_0\ldots x_d)$ for credibility evaluation $(\textrm{Pr}(x_0|x_1,\ldots,x_d))$ by just substituting $x_1\ldots x_d$ and normalizing obtained polynomial of $x_0$ to integrate to 1. This way $f_{j_0}(x_0)$ is expressed as nearly a linear combination of various products of $f_{j_k}(x_k)$. However, this approach has difficulties with discrete values - hence, there is finally used linear regression to directly optimize coefficients of such linear combinations.\\

Orthogonal polynomial basis allows also for a different perspective: for a given $x\in[0,1]$ value, we can take some its coordinates in this basis: $(f_j(x))_{j=1\ldots m}$. As estimated $a_j$ is just average over such corresponding coordinates, we can imagine e.g. $f_1(x)$ as contribution of this point to expected value, $f_2(x)$ to variance etc. Hence, we can use $(f_j(x))_{j=1\ldots m}$ as a list of features of a given data point for inference of its other features, for $j\geq 2$ exploiting nonlinear dependencies this way - we will use such features in linear regression for predictions here.

There is a large freedom of choice for features of variables to infer from in prediction. For example for age variable here, a standard approach is dividing into age ranges, what can be seen as using $(f_j(x))_j$ with family of functions being 1 on a given age range and 0 otherwise. It leaves a question of sizes of these age ranges: long ranges reduce data precision, short ranges mean lower statistics and not exploiting local behavior (of neighboring ranges). Normalizing variable and using $(f_j(x))_j$ with orthonormal family of polynomials is an attempt to systematically exploit local dependencies, making each coefficient being moment-like and affected by all data points. Numerical tests for age gave slightly more accurate predictions using orthonormal polynomials, than using the same number of fixed length age ranges.
\subsection{Handling discrete variables}
Analogously to continuous variables with $\langle f,g\rangle =\int_0^1 f(x)g(x) dx$ scalar product, we could directly work on discrete variables with $[ f,g ]=\sum_x w_x f_x g_x$ scalar product using some weight e.g. chosen as marginal frequencies $w_x=g_x$ for $g_x=\textrm{Pr}(x^i=x)=|\{i:x^i=x\}|/n$ in analogy to continuous weights. Using orthonormal basis $[ f_j, f_k]=\delta_{jk}$, we get analogous MSE estimation as $a_j =[f_j,g]$.

The basis should analogously start with $f_0=g$ marginal distribution, then further vectors describe distortion from it. Orthonormality makes $f_1$ unique for binary alphabet. For a lager one we can choose initial vectors using orthonormal polynomials on variables normalized with $(\ref{norm})$, then perform Gram-Schmidt orthonormalization using $[\cdot,\cdot]$ scalar product. To simplify and to present intuition of treating $(f_j(x))_j$ as useful features for inference, which can be imagined as $x$-th contribution to $j$-th moment, presented analysis instead uses linear regression to directly optimize coefficients.
\section{Used algorithm}
\begin{figure*}[b!]
    \centering
        \includegraphics{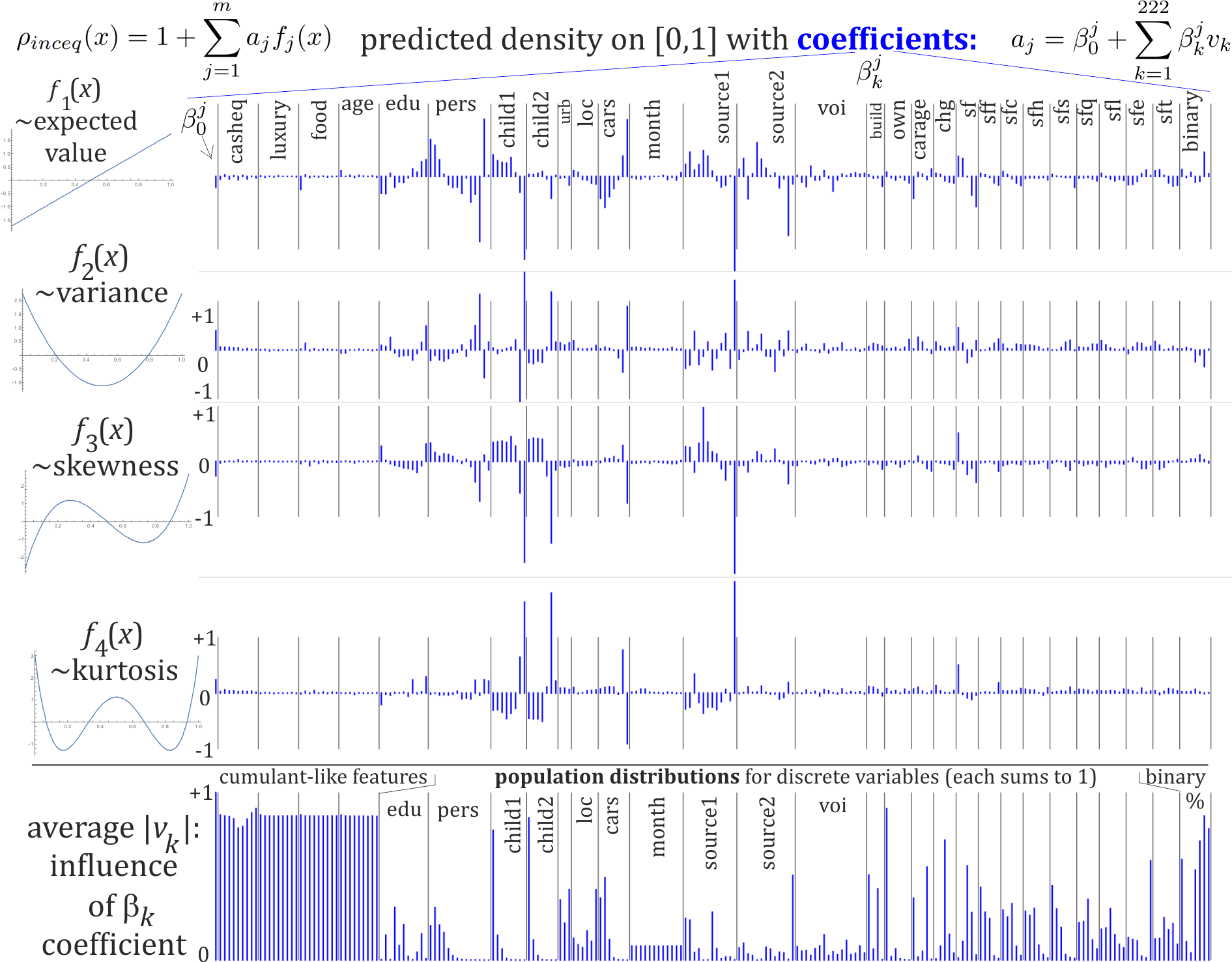}
        \caption{Top: final $4\times 223$ coefficients obtained by least-square regression (optimizing $\sum_i \|f_j(x^i_0)-a_j\|^2$) for predicting probability density of exogenous variable (equivalent income) based on features of endogenous variables - independently for coefficients corresponding to expected value $(f_1)$, variance $(f_2)$, skewness $(f_3)$ and kurtosis $(f_4)$ of predicted normalized variable $x_0$.
        For endogenous variables treated as continuous (casheq, luxury, food, age), the used features are $f_j(x_l)$ for $j=1\ldots 9$ and $l=1,2,3,4$ describing $j$-th cumulant-like behavior of $l$-th variable. The remaining variables are binary or discrete treated as binary: 0 or 1 for each appearing possibility. For example in the first row, negative first coefficient for food connects their expected values - describes anticorrelation, analogously for age it is positive - they are correlated. Coefficients for voivodeships (voi) describe individual corrections for each geographical region. Low statistics for some coefficients can lead to surprising behavior, for example we can see reduction of equivalent income with the number of persons in the household (pers), with a surprising spike at the end - it corresponds to a single data point of 12 person household.
        Bottom: average $|v_k|$ describing average influence of $\beta_k$ coefficient - it is fixed 1 for $v_0$, large nearly constant for cumulant-like coefficients of  continuous variables (increasing their importance), for each binarized discrete variable its contributions sum to 1: they describe proportions of such categories in the population, for binary variables each is in $[0,1]$.  }
       \label{coef}
\end{figure*}

The currently used algorithm optimizes coefficients with least-square regression:
\begin{enumerate}
  \item All variables treated as continuous - including casheq having a large percentage of exactly 0 value, and age obtaining 87 distinct discrete values - are  normalized to nearly uniform marginal distribution using formula (\ref{norm}).
  \item All the remaining variables are treated as categorical and transformed into binary - thanks of it, weights of individual categories are optimized in later regression. For example edu(cation) obtains 11 different values, hence it is transformed into 11 binary variables: each being 1 if category agrees, 0 otherwise.
  \item Denote $v(y_1\ldots y_d)$ as vector built of all features of endogenous variables (normalized or not) directly used for prediction as a linear combination - here it has 223 coordinates visualized in Fig. \ref{coef}. Its zeroth coordinate is fixed as 1 to get constant term $(\beta_0)$ in later regression. Then for variables treated as continuous, it contains $f_j(x_k)$ for $j=1$ up to a chosen degree, which is 9 here for all 4 variables treated this way (casheq, luxury, food, age) - getting $4\times 9 =36$ coordinates of $v$. It further contains all the remaining variables - categorical transformed into binary, and the original binary variables - both using only 0 or 1 values.
  \item Build e.g. $n\times 223$ matrix $M$ with rows being applied  $v(y^i_1\ldots y^i_d)$ function to all $i=1\ldots n$ data points.
  \item We would like to use least square linear regression to infer $f_{j}(x_0)$ for $j=1\ldots m$ from $v(y_1\ldots y_d)$ to predict density as degree $m$ polynomial. For this purpose,  build vectors $b^j=(f_{j}(x^i_0))_{i=1\ldots n}$, then find coefficient vectors: $\beta^{j}$ minimizing $\|M\beta^j-b^j\|$. It can be realized with pseudoinverse, and is implemented in many numerical libraries, e.g. as "LeastSquares[M,b]" in Wolfram Mathematica. Values of these final used coefficients are visualized in Fig. \ref{coef}.
  \item Now predicted density is $$\rho_0(x_0)=1+\sum_{j=1}^m a_j f_j(x_0)\quad\textrm{for}\quad a_j=v(y_1\ldots y_d)\cdot \beta^j$$
\end{enumerate}
This predicted density is for exogenous variable normalized to nearly uniform marginal distribution on $[0,1]$ - what allows to use it directly for credibility evaluation.

\begin{figure}[t!]
    \centering
        \includegraphics{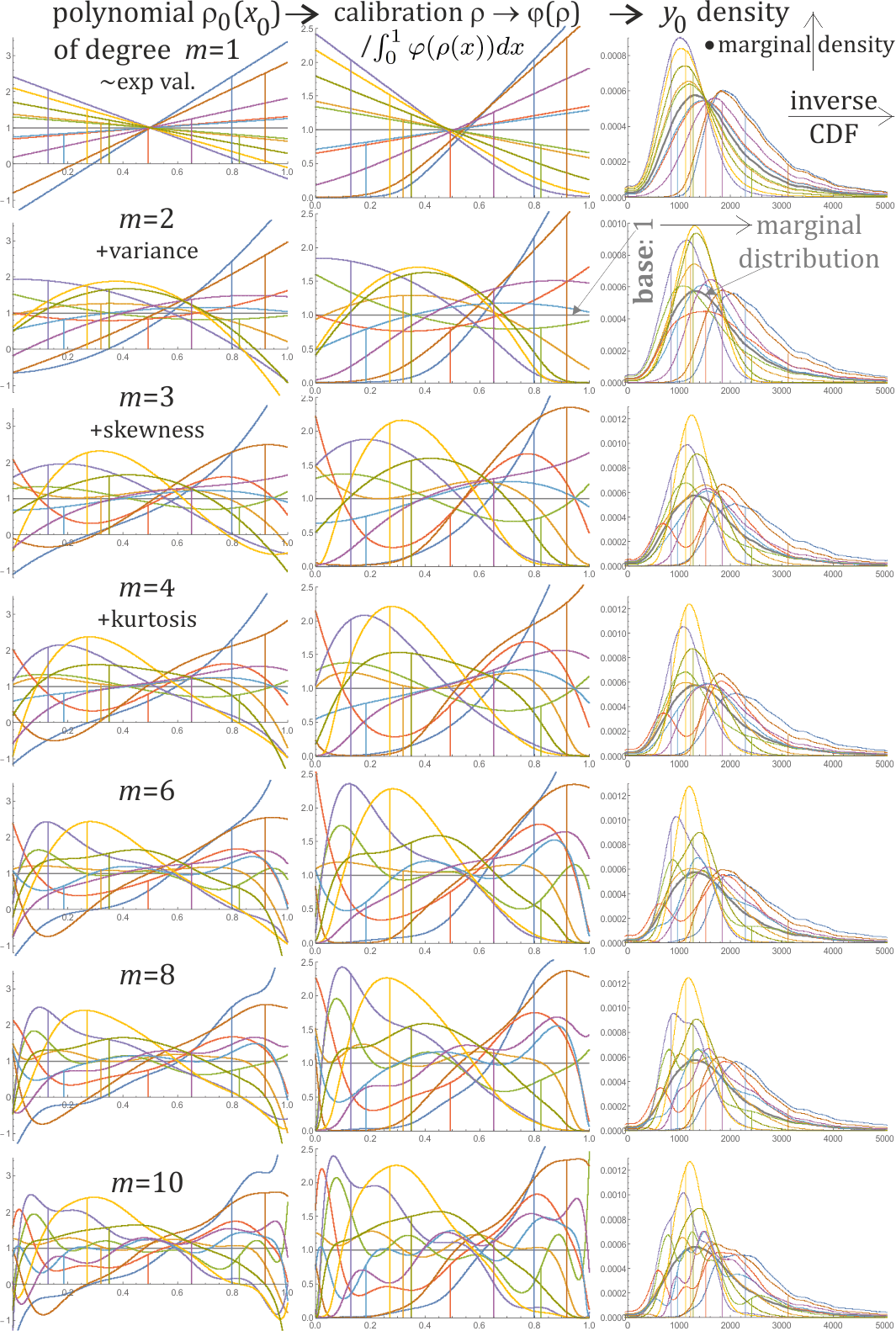}
        \caption{Predictions of density for inceq based on the remaining variables for 10 randomly chosen points (colors, the same in all plots). Row corresponds to used degree $m$ polynomial predicting density (as in Fig. \ref{intr}) - they are presented in the left column, together with the actual values as vertical lines of the same color. Middle column: calibration as discussed in \cite{me3} to actual nonnegative densities from such predictions, using function $\varphi(\rho)=\ln(1+\exp(5\rho)/2)/5$ here, which was obtained by approximating division of density of actual values by density of all predicted values. Beside applying $\rho\to\varphi(\rho)$ we need to divide by $\int_0^1 \varphi(\rho(x))dx$ to ensure integrating to 1. Right column: translating to predicted density of original variable ($x\to y$, removing normalization with EDF) - position on horizontal axis is obtained by using $CDF^{-1}$, on vertical by multiplying calibrated density by marginal density for a given point - presented as thick gray line, obtained from $\rho=1$ base density while using no prediction.}
       \label{poly}
\end{figure}

Sorting such predicted densities of actual values, we can find thresholds for some percentage of least credible elements, for example for $\approx 1\%$ of population this threshold here is 0: this way obtaining negative values of predicted polynomials can be treated e.g. as requiring some additional verification.

If required we can also transform such polynomials into real predicted probability distributions by using some calibration function $\rho=\varphi(\rho_0)>0$ and normalizing to integrate to 1, like presented in fig. \ref{poly}. A simpler choice is just $\varphi(\rho)=c\, \max(\rho,\epsilon)$ for some $\epsilon>0$ and $0<c\leq 1$ calculated to integrate to 1. This figure also shows translating such predicted distribution from $x\in [0,1]$ into the original variable $y$: by multiplying the marginal distribution by predicted density in corresponding positions.

Such distributions can be used various purposes. For example if we are only interested in predicting value, we can calculate expected value of such final prediction. Doing it through normalization to $[0,1]$ here repairs instability problems of predicting tail behavior by standard regression. To calculate such expected value we can take regular lattice on $[0,1]$ of size being the number of points, calculate polynomial value on this lattice, apply calibration function, divide by sum to represent discrete probability distribution, and treat them as probabilities of sorted values to calculate expected value:
$$\sum_{i=1}^n\, y_0^{o(i)}\, \varphi(\rho_0((i-0.5)/n))\, /\sum_{j=1}^n \varphi(\rho_0((j-0.5)/n))$$

\begin{figure}[t!]
    \centering
        \includegraphics{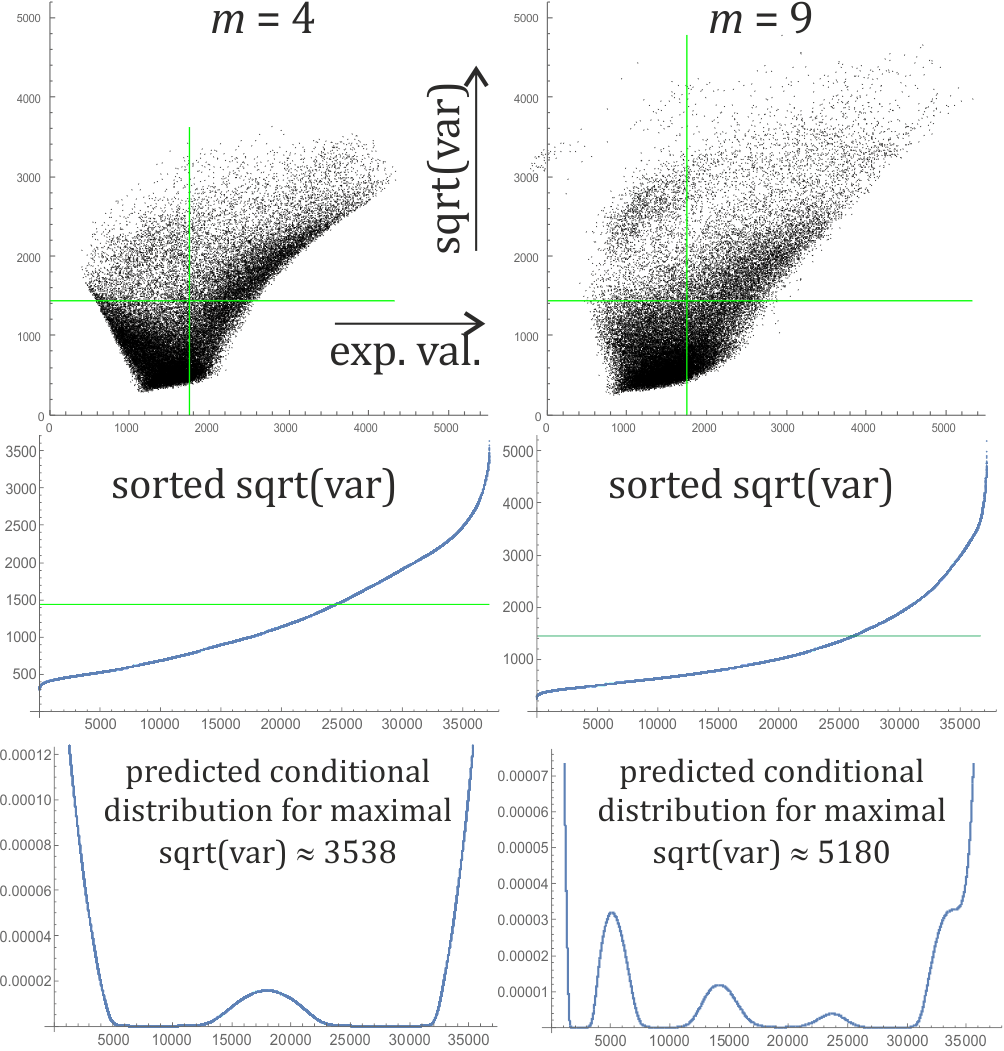}
        \caption{Top: pairs of final (predicted expected value, square root of variance) for all data points and $m=4$ (left) or $m=9$ (right) used polynomial degree. The green lines are expected value and square root of variance of the dataset. We can see that predictions can give larger variance than of dataset, especially for large degree $m$, what means strengthened tails in predicted often multimodal densities. Middle: sorted squared roots of variances of predicted distributions, we can see that $\approx 1/3$ are above square root of variance of the sample. Bottom: predicted conditional distributions  having largest variance in both cases, both have densities concentrated in two tails. }
       \label{expvar}
\end{figure}

We can analogously calculate e.g. variance (or higher moments) - estimating uncertainty of predicted expected value. Pairs (expected value, variance) calculated this way  for discussed data are presented in Fig. \ref{expvar}.

\section{Conclusions and further work}
There was presented a simple general approach for applying HCR methodology with discrete variables and alternative: direct estimation of parameters - on example of credibility evaluation for income data, allowing to model conditional probability distribution for the predicted variable.

It tests agreement with statistics of provided data sample - will not detect systematic improper behavior existing in this sample. Handling it would rather require some supervision, like manually marking some of suspicious data points and then evaluate probability of being in this marked set.\\

The main purpose of this article was methodological - presenting simple interpretable analysis, but leaving many possibilities for improvements.

For example it has exploited only pairwise dependencies - between the exogenous variable and (separately) each of endogenous variable. We can analogously include higher order dependencies by using products of considered coordinates for vector $v$ of above algorithm, e.g. include $f_{j_1}(x_1)f_{j_2}(x_2)$ features in the least square regression for 3-point correlation, and analogously for higher order dependencies. As number of possibilities grows exponentially with dependency order, this choice requires some selectivity - for example can be done in hierarchical way: e.g. search for non-negligible 3-point correlations by expanding essential 2-point correlations. Another option is using $L_1$ regularization (called lasso method): add $\sum_\textbf{j} |a_\textbf{j}|$ to optimized criterion, leading to a sparser basis selected by non-negligible coefficients. An inexpensive way of selecting coefficients to be included is taking those of large absolute values. A better way is testing if such additional coefficients improve log-likelihood of prediction, using separate training and evaluation set like for table in Fig. \ref{intr}.

The main purpose of using least-square regression was simplicity, interpretability and presenting perspective of moment-like features. It is a bit different approach than standard HCR, still leading to fruitful predictions - their relation needs to be further investigated.

As mentioned, another direction worth exploring is using dimensionality reduction techniques like PCA or its discrete analogue: MCA, to reduce the number of variables and make them continuous.
% $$\rho_{inceq}(x)=1+\sum_{j=1}^m a_j f_j(x) \qquad a_j=\beta^j_0 + \sum_{k=1}^{222} \beta_{k}^j v_k $$

\bibliographystyle{IEEEtran}
\bibliography{cites}
\end{document}